\def\year{2018}\relax
\documentclass[letterpaper]{article} 
\usepackage{aaai18}  
\usepackage{times}  
\usepackage{helvet}  
\usepackage{courier}  
\usepackage{url}  
\usepackage{graphicx}  
\begin{document}
%
\title{Clustering - What Both Theoreticians and Practitioners are Doing Wrong{}}
\author{Shai Ben-David\\
Cheriton School of Computer Science\\
University of Waterloo\\
Waterloo, Canada\\
}
\maketitle
\begin{abstract}
Unsupervised learning is widely recognized as one of the most important challenges facing machine learning nowadays. However, in spite of hundreds of papers on the topic being published every year, current theoretical understanding and practical implementations of such tasks, in particular of clustering, is very rudimentary.

This note focuses on clustering. I claim that the most significant challenge for clustering is model selection. In contrast with other common computational tasks, for clustering, different algorithms often yield drastically different outcomes. Therefore, the choice of a clustering algorithm, and their parameters (like the number of clusters) may play a crucial role in the usefulness of an output clustering solution. 

However, currently there exists no methodical guidance for clustering tool-selection for a given clustering task. Practitioners pick the algorithms they use without awareness to the implications of their choices and the vast majority of theory of clustering papers focus on providing savings to the resources needed to solve optimization problems that arise from picking some concrete clustering objective. Saving that pale in comparison to the costs of mismatch between those objectives and the intended use of clustering results. I argue the severity of this problem and describe some recent proposals aiming to address this crucial lacuna.
\end{abstract}

\section{Introduction}

Clustering is one of the most basic and useful data processing tasks. It is being routinely applied in a wide variety of applications.
Not surprisingly, there exist many clustering algorithms. However, clustering
is an ill defined problem - given a data set, it is not clear what a \emph{correct}
clustering for that set is. Indeed, different algorithms may yield dramatically
different outputs for the same input sets. In contrast with other common learn-
ing tasks, like classification prediction, clustering does not have a well defined
ground truth. Faced with a concrete clustering task, a user needs to choose an
appropriate clustering algorithm (as well as a concrete setting for the tuneable
parameters of the chosen algorithm). Currently, such decisions are often made
in a very ad hoc, if not completely random, manner. Given the crucial effect
of the choice of a clustering algorithm on the resulting clustering, this state of
affairs is truly regrettable. Can the research community develop effective tools
for helping users make informed decisions when they come to pick a clustering
tool for their data? How can we help the data analysts in finding the cluster
structures that will best suit their needs?

Here, we focus on the basic case, where the relevant similarity (or distance measure) is given.
Furthermore, in this note, we only consider partitional deterministic clustering, where each data point ends up in single cluster. 
Most of the issues that we raise are relevant to other clustering settings as well, however, for the sake of concreteness we keep that outside the scope of the current discussion.
Finally, it is worthwhile mentioning that we consider the aspect of clustering that aims to group data in a way that similar points share a cluster and dissimilar points are separated into different clusters. The term clustering is sometimes also applied to the 
detection of connected components of density level sets of a probability distribution. This is a somewhat different interpretation of clustering and will not be addressed here.

\section{The conflicts between various desiderata}
Arguably the most basic definition of clustering is \emph{``partitioning of data into groups (a.k.a. clusters) so that similar (or close w.r.t. the underlying distance function) elements share the same cluster and the members of each cluster are all similar"} (or, equivalently, dissimilar elements are separated into different clusters). A moment reflection reveals that this definition is problematic. Its two requirements may well conflict with each other. As a simple example, consider a collection of elements scattered next to each other along a long line. If we wish to satisfy ``every pair of close by elements share the same cluster" we ought to put all of those points in a single joint cluster. However, such a clustering violates the second requirement - we will end up with dissimilar elements sharing the same cluster. If you wish, this is the simplest \emph{clustering impossibility theorem}. More abstractly, the goal of partitional clustering is to have clusters such the relation "$x$ and $y$ belong to the same cluster" is a good approximation to the input relation "$x$ is similar to $y$". However, the first relation is transitive whereas the relation "being similar" may well violate transitivity. 

 The above basic definition fails to determine how should such conflicts be resolved. Furthermore, this is not the only potential conflict.
 There are many other clustering properties that are desirable under some conceivable circumstances. For example, keeping some
balance between the number of elements in different clusters, or being robust to small data perturbations, and more. For any pair of such requirements there are data sets for which those requirements cannot be mutually met.

Indeed, as we will elaborate in the next sections, different clustering paradigms prioritize those requirements differently, and different clustering applications call for different prioritization. 

\subsection{Some common clustering approaches emphasize different requirements}
Let us briefly consider some basic clustering paradigms from the point of view of the above desiderata;
\begin{itemize}
\item Its is easy to see that Single Linkage (SL) clustering (with some fixed termination rule) focuses on the requirement
``every pair of close by elements share the same cluster". If points are close enough (relative to the set of point wise distances in the data), SL will keep them in the same cluster. However, SL is oblivious to generating clusters that contain very far apart points or to extreme imbalance between the sizes of the clusters it ends up with.
\item Similarly, Max Linkage focus on the requirement ``the members of each cluster are all similar". For some data sets, it outputs clusters that separate very (relatively) similar points, and it is oblivious to imbalance between the sizes of the clusters it generates.
\item In contrast with both of the above paradigms, as well as with any other linkage based clustering rule,  the $K$-Means algorithm is sensitive to imbalance between the number of points in different clusters. It may well end up
with clusters that cut through a dense cloud of points while including far away elements, for the sake of saving the cost incurred  by having too many points in one cluster.

\end{itemize}

Similarly, one can readily realize that other clustering paradigms prioritize different requirements differently.

\subsection{Some common clustering tasks call for different prioritizations of clustering properties}
Here we give just a few generic examples of common usages of clustering that obviously call for taking into consideration different clustering requirements:

\begin{itemize}
\item \emph{Clustering records in a large data base to eliminate duplications} (say records of patients collected from many clinics and hospitals to detect records that refer to the same patient). Clearly, a high priority for such an application of clustering is not having very different elements (records) share the same cluster (namely, be declared as referring to the same patient).

\item \emph{Clustering natural vegetation for predicting the potential spread of some fungus}.  In such cases, the foremost consideration in determining clusters should be that close by plants should belong to the same cluster, since the fungus is likely
to pass between them.

\item \emph{Clustering neighbourhoods into school districts}. For such an application the requirement of (some) balance 
between the sizes of different clusters should be an important consideration.

\end{itemize}

The conclusion is obvious - different clustering tasks should be carried out by different clustering algorithms, and that choice
of clustering tools should be done based on careful examination of the clustering task as well as elaborate understanding of the 
features of each clustering algorithm (and often also the effect of picking appropriate parameter values for those algorithms).

\section{What should practitionares change?}

When you ask clustering users how did they pick the specific clustering algorithm they are using on their data,
the answers one gets sound extremely ad hoc: ``This is the algorithm that is commonly used in my field", ``This algorithm is 
user friendly and easy to run", ``This algorithm is provided with the statistics package I got", ``This algorithm runs fast", ``no need to tune any sensitive parameters". 

I hear such answers even from the most otherwise-sophisticated users, biologists, astronomers, physicists, social scientists,
you name it.

For analogy, assume you are sick, your doctor prescribes some medication for you and you ask "Why did you pick this particular medication?". Will you accept answers like "This is what my peers like to use" (regardless of what your disease is), or ``This medication
can be taken in any time of the day, regardless of when you eat", or ``This medication is the cheapest drug in the market"?
Will you pick an over the counter drug just based on the aesthetics of its cover, or its price, without reading the details of what it does and for which symptoms should it be used?

As silly as it sounds, the way many users pick their clustering tool is not much more thoughtful.

\section{What should theoreticians change?}
Every big data science related conference has many clustering papers (last year's AAAI had about 20 of those). 
What do most of those papers offer? They talk about run time, they give examples of data sets on which the results were 
great, they may offer a new algorithmic approach. However, few, if any at all, make an attempt to analyze with respect to which clustering properties do their algorithms differ from existing ones? For which specific applications will the proposed algorithm be more suitable and why? 

For the sake of saving face, I refrain for mentioning any specific papers. But just check it out -- those issues are very common. In AAAI papers, in NIPS papers, in Science and Nature papers, all around us.

Furthermore, in the algorithms and complexity research communities, a lot of effort is devoted to coming up
with efficient approximation algorithms for clustering objective minimization (such as approximating  $k$-means cost minimization
with runtime that is not exponential in the number of clusters, $k$).
I would argue that in most clustering applications the users do not care about finding such a cost minimizing solution. In fact,
they can never even evaluate how close to the minimum possible is the cost of the clustering output by some algorithm.
A suitable choice of an algorithm (maybe through a good choice of an objective function) will usually have a much higher impact on the usability of output clusterings. Just the same, very little, if any, research is devoted to that algorithmic selection aspect.

\section{Recent work on informed choice of clustering tools}
The challenge of matching clustering tools to clustering tasks has been addressed along two lines of research.
\subsection{An analytic approach}
A natural option is the establish a list of properties, or features, of clustering algorithms that would serve as a basis for 
informed search for users faced with a concrete clustering task. Ideally, based on such a list clustering algorithms will each have a profile - a vector of scores of their degree of compliance with each of these properties. Such an approach has been proposed, for example, by Ackerman et al \cite{rita-nips2}.

The main challenge with this direction is finding a way of expressing properties of clustering algorithms that is meaningful at the same time both for algorithm design purposes and to clustering end users. One issue with the properties proposed so far (e.g. by \cite{rita-nips2} and reference there) is that they address the behaviour of an algorithm over \emph{all possible} data sets, or with respect to some statistics over data sets, while a user is concerned with their own specific data. In some sense, such ``worst case'' or ``average case'' behaviour may not be relevant to a given subset of data sets of interest. An algorithm may fail to satisfy some such requirement that a user cares about on some possible inputs and yet meet that same requirement on the data the user cares about.

\subsection{Interactive semi-supervised clustering}
An alternative approach is to rely on interactions between the data expert and the algorithm designer. This may be viewed as 
providing some supervision (on top of raw, unlabeled data) to the clustering procedure, hence the term Semi-Supervised Clustering.
This approach dates back to \cite{Wagstaff00} (see a survey of such methods in \cite{Blair_survey}).

Statistical analysis of a specific type of semi-supervision, a model in which the user provides a desired clustering of a small sample of the given data set, is provided in \cite{Hassan_uai}.

Another component allowing introducing domain specific bias into a clustering tool is ongoing interaction between the user and the algorithm as modelled by interactive clustering, e.g.,  \cite{Nina_2017} and references there within.

\cite{Hassan_Nips2016} demonstrates how allowing active pairwise sam-cluster queries not only allows domain knowledge to effect the 
output clustering but may also result in dramatic reduction of the clustering algorithm's run time.

\section{Challenges and directions for further research}
The first message I would like to convey is the importance of researching and developing tools for the transfer of domain expertise
and task requirements between the end user of clustering and the algorithm developer. 
First steps in this direction are being slowly taken, but much more is needed.

Another important message concerns the way clustering research is presented. Researchers should pay much more attention to 
understanding and explaining the elaborate combination of requirements that distinguishes the methods they propose, and to the range of tasks they are good for. There can be no universally optimal clustering algorithm. Your algorithm is never just "better" than any competitor's algorithm, it may only be better for some specific clustering tasks, try to understand and explain what are those.

Let me mention just one of the many other shortcomings of our current understanding of clustering. When faced with a task and data  that is likely to result in a large number of clusters (like the record de-duplication task), the user cannot tell in advance what the number of resulting clusters should be. Furthermore, there is no clear monotonicity - we would not necessarily be happier with clustering with more clusters or with clustering with less clusters. For such tasks, we currently have no satisfactory objective (a.k.a. cost) function
to guide the a clustering algorithms. 

\bibliographystyle{aaai}
\bibliography{refs}
\end{document}